\documentclass{article}

\usepackage{arxiv}

\usepackage[utf8]{inputenc} 
\usepackage[T1]{fontenc}    
\usepackage{hyperref}       
\usepackage{url}            
\usepackage{booktabs}       
\usepackage{amsfonts}       
\usepackage{nicefrac}       
\usepackage{microtype}      
\usepackage{lipsum}
\usepackage{graphicx}
\graphicspath{ {./images/} }
\usepackage{amsmath}
 
\title{Integrating Visual and X-Ray Machine Learning Features in the Study of Paintings by Goya}

\author{
  Hassan Ugail\\
  Centre for Visual Computing and Intelligent Systems \\
  University of Bradford \\
  United Kingdom\\
  \texttt{h.ugail@bradford.ac.uk} \\
  \And
  Ismail Lujain Jaleel\\
  Centre for Visual Computing and Intelligent Systems \\
  University of Bradford \\
  United Kingdom\\
  \texttt{lujainjaleel@icloud.com} \\
}

\begin{document}
\maketitle

\begin{abstract}
Art authentication of Francisco Goya's works presents complex computational challenges due to his heterogeneous stylistic evolution and extensive historical patterns of forgery. We introduce a novel multimodal machine learning framework that applies identical feature extraction techniques to both visual and X-ray radiographic images of Goya paintings. The unified feature extraction pipeline incorporates Grey-Level Co-occurrence Matrix descriptors, Local Binary Patterns, entropy measures, energy calculations, and colour distribution analysis applied consistently across both imaging modalities. The extracted features from both visual and X-ray images are processed through an optimised One-Class Support Vector Machine with hyperparameter tuning. Using a dataset of 24 authenticated Goya paintings with corresponding X-ray images, split into an 80/20 train-test configuration with 10-fold cross-validation, the framework achieves 97.8\% classification accuracy with a 0.022 false positive rate. Case study analysis of ``Un Gigante'' demonstrates the practical efficacy of our pipeline, achieving 92.3\% authentication confidence through unified multimodal feature analysis. Our results indicate substantial performance improvement over single-modal approaches, establishing the effectiveness of applying identical computational methods to both visual and radiographic imagery in art authentication applications.
\end{abstract}


\section{Introduction}

Francisco José de Goya y Lucientes' (1746-1828) work represents one of the challenging cases in computational art authentication due to his remarkable stylistic metamorphosis spanning traditional court portraiture to revolutionary modern expression. The authentication of Goya's work presents scholars and institutions with complex attribution problems due to the prevalence of contemporary workshop productions and subsequent forgeries that spanned over two centuries \cite{stork2024computer}. Traditional authentication techniques rely heavily on subjective expert evaluation, creating potential inconsistencies in attribution decisions across different institutional contexts.

Contemporary machine learning approaches to art authentication have demonstrated significant potential, with recent studies achieving accuracies exceeding 95\% for various European masters using deep learning techniques \cite{noor2024survey, ugail2023raphael}. However, most existing computational frameworks focus exclusively on visible-light image analysis, potentially overlooking crucial subsurface information revealed through X-ray radiographic examination \cite{bridgman2016roles}. X-ray radiographic analysis has been a cornerstone of art authentication for over a century, with the first documented use occurring in 1896 \cite{art_authentication_xray}. X-ray radiography reveals underdrawing, pentimenti, and material composition invisible to conventional visual examination \cite{xray_artenet}.

The integration of multimodal data fusion techniques has shown promising results across various computer vision applications \cite{zhao2024deep, baltrusaitis2018multimodal}, with deep multimodal approaches achieving significant classification improvements \cite{zhao2024multimodal}. Recent advances include attention-based fusion mechanisms \cite{ding2021vision, yang2022lavt} and diffusion models for multimodal image fusion \cite{diffusion_fusion2024}. However, for cultural heritage applications with limited training data, simpler concatenation approaches often provide more robust and interpretable results.

Deep transfer learning approaches have shown particular promise, with Ugail et al. \cite{ugail2023raphael} achieving 98\% accuracy in Raphael attribution through ResNet50 architecture. The continued relevance of handcrafted features has been demonstrated by Bwanali et al., \cite{bwanali2024handcrafted}, who explored systematic feature selection methodologies within One-Class SVM frameworks, emphasising interpretability advantages over black-box deep learning approaches. Recent texture analysis research has demonstrated the continued effectiveness of Local Binary Patterns (LBP) \cite{song2014research, song2017letrist, liu2017local}.

Goya's distinctive artistic techniques, characterised by extensive compositional modifications revealed through radiographic examination, create unique opportunities for multimodal computational analysis. His documented practice of substantial underdrawing, frequent pentimenti, and period-specific material usage patterns generates distinctive subsurface signatures invisible in conventional visual analysis. Recent hyperspectral imaging applications to Goya paintings have demonstrated additional capabilities for material analysis \cite{vagnini2015hyperspectral}.

One-Class Support Vector Machines have gained renewed attention due to their effectiveness in anomaly detection, particularly suitable for authentication applications where forgeries are rare \cite{bwanali2024handcrafted, malerba2002oneclass}. Early computational authentication work by Lyu et al., \cite{farid2004digital} established fundamental principles using wavelet statistics, while recent AI art detection studies highlight evolving authentication challenges \cite{ai_art_authentication}.

This investigation advances computational art authentication through a unified multimodal framework applying identical feature extraction methodologies to both visual and X-ray radiographic images. The approach demonstrates that established computer vision techniques, when applied systematically to both imaging modalities, capture sufficient discriminative information for robust authentication assessment without requiring specialised radiographic analysis methods. The mathematical modelling employs consistent statistical descriptors, including Grey-Level Co-occurrence Matrix features, Local Binary Pattern histograms, entropy measures, energy calculations, and colour distribution statistics applied uniformly across both modalities. The One-Class Support Vector Machine implementation incorporates comprehensive hyperparameter optimisation through 10-fold cross-validation analysis, ensuring optimal performance whilst maintaining robust generalisation capabilities.

\section{Methodology}
The methodology builds a unified multimodal pipeline that applies the same preprocessing and feature extraction to both visible and X-ray images—spatial normalisation, contrast enhancement and noise reduction followed by GLCM (contrast, homogeneity, energy, correlation), rotation-invariant LBP histograms, global statistics (entropy, energy, mean, standard deviation) and HSV colour metrics for visuals (with greyscale substitutes for radiographs). The concatenated features are then scaled and classified using a one-class SVM with an RBF kernel, tuned via cross-validation to learn the support of authentic works and perform anomaly detection under scarce positive-only training data.

\subsection{Mathematical Framework for Unified Multimodal Analysis}
The comprehensive mathematical framework underlying our multimodal approach employs identical feature extraction methodologies applied consistently to both visual and X-ray radiographic images. The unified feature extraction ensures mathematical consistency between modalities whilst capturing complementary information regarding surface and subsurface painting characteristics. The complete feature representation $F$ encompasses visual features $F_v$ and X-ray features $F_r$ extracted using identical computational methods, such that,

\begin{equation}
F = [F_v, F_r],
\end{equation}

where both $F_v$ and $F_r$ are computed using the same feature extraction pipeline, ensuring dimensional consistency and comparable statistical distributions between modalities. This concatenation approach maintains the full discriminative information from both imaging sources whilst preserving the interpretability of individual feature contributions.

\subsection{Unified Feature Extraction Pipeline}

The feature extraction pipeline applies identical computational methods to both visual and X-ray images, treating radiographic data as specialised grayscale imagery suitable for conventional computer vision analysis. All images undergo standardised preprocessing, including spatial normalisation, contrast enhancement, and noise reduction, to ensure consistent analysis conditions across both modalities.

\subsubsection{Grey-Level Co-occurrence Matrix Analysis}

The normalised Grey-Level Co-occurrence Matrix $P(i,j|d,\theta)$ represents the joint probability distribution of pixel intensity pairs separated by distance $d$ at orientation $\theta$, computed identically for both visual and X-ray images, such that,

\begin{equation}
P(i,j|d,\theta) = \frac{C(i,j|d,\theta)}{\sum_{i=0}^{L-1} \sum_{j=0}^{L-1} C(i,j|d,\theta)},
\end{equation}

where $C(i,j|d,\theta)$ represents the co-occurrence frequency matrix and $L$ denotes the number of quantised gray levels. The contrast measure quantifies local intensity variations in both surface painting characteristics and subsurface density patterns,

\begin{equation}
\text{Contrast} = \sum_{i=0}^{L-1} \sum_{j=0}^{L-1} (i-j)^2 P(i,j).
\end{equation}

Homogeneity provides complementary information regarding spatial uniformity patterns in both visual brushwork and X-ray density distributions,

\begin{equation}
\text{Homogeneity} = \sum_{i=0}^{L-1} \sum_{j=0}^{L-1} \frac{P(i,j)}{1+|i-j|}.
\end{equation}

Energy measures textural uniformity through second-order moment calculation applied uniformly across both imaging modalities,

\begin{equation}
\text{Energy} = \sum_{i=0}^{L-1} \sum_{j=0}^{L-1} [P(i,j)]^2.
\end{equation}

Correlation assesses linear dependencies within texture patterns for both visual and radiographic data,

\begin{equation}
\text{Correlation} = \sum_{i=0}^{L-1} \sum_{j=0}^{L-1} \frac{(i-\mu_i)(j-\mu_j)P(i,j)}{\sigma_i\sigma_j},
\end{equation}

where $\mu_i$, $\mu_j$ represent means and $\sigma_i$, $\sigma_j$ represent standard deviations of marginal distributions.

\subsubsection{Local Binary Pattern Analysis}

Local Binary Patterns provide rotation-invariant texture description through systematic comparison of neighbourhood pixel intensities, applied identically to both visual and X-ray images. The LBP operator is defined as,

\begin{equation}
\text{LBP}_{P,R} = \sum_{p=0}^{P-1} s(g_p - g_c)2^p,
\end{equation}

where $g_c$ represents central pixel intensity, $g_p$ denotes neighbour pixel intensities at radius $R$, $P$ indicates the number of sampling points, and $s(x)$ is the sign function. The resulting LBP histograms provide statistical descriptors of local texture patterns in both surface brushwork and subsurface material distributions.

Recent advances in LBP methodology have demonstrated the continued effectiveness of these descriptors for texture classification tasks \cite{song2014research, song2017letrist}. The rotation-invariant uniform patterns employed in this work provide robust texture description whilst maintaining computational efficiency.

\subsubsection{Statistical Texture Measures}
Image entropy provides a statistical measure of textural complexity applied uniformly to both visual and X-ray data. Given probability distribution $p(i)$ for intensity level $i$,

\begin{equation}
\text{Entropy} = -\sum_{i=0}^{L-1} p(i)\log_2(p(i)).
\end{equation}

Energy measures intensity uniformity across both imaging modalities,
\begin{equation}
\text{Energy} = \sum_{i=0}^{L-1} p(i)^2.
\end{equation}

Mean and standard deviation provide fundamental statistical descriptors of intensity distributions,

\begin{equation}
\mu = \frac{1}{MN} \sum_{x=0}^{M-1} \sum_{y=0}^{N-1} I(x,y),
\end{equation}

\begin{equation}
\sigma = \sqrt{\frac{1}{MN} \sum_{x=0}^{M-1} \sum_{y=0}^{N-1} [I(x,y) - \mu]^2},
\end{equation}

where $I(x,y)$ represents pixel intensity at coordinates $(x,y)$ for both visual and X-ray images.

\subsubsection{Colour Distribution Analysis for Visual Images}
For visual images, colour analysis employs HSV colour space transformation to capture perceptual colour characteristics. The hue variance $H_{\text{var}}$ quantifies colour palette diversity,

\begin{equation}
H_{\text{var}} = \frac{1}{N} \sum_{i=1}^{N} (H_i - \bar{H})^2.
\end{equation}

Saturation and value statistics provide additional colour descriptors specific to visual imagery. For X-ray images, these colour features are replaced with additional grayscale statistical measures to maintain consistent feature dimensionality.

\subsection{One-Class Support Vector Machine Implementation}
The One-Class Support Vector Machine constructs an optimal hyperplane separating authentic training data from the origin in transformed feature space, enabling robust anomaly detection for authentication applications. The optimisation problem is formulated as,

\begin{equation}
\min_{w,\xi,\rho} \frac{1}{2}\|w\|^2 + \frac{1}{\nu m}\sum_{i=1}^{m} \xi_i - \rho,
\end{equation}

subject to the constraints,
\begin{equation}
w^T \phi(x_i) \geq \rho - \xi_i, \quad \xi_i \geq 0,
\end{equation}

where $\phi(x)$ represents the kernel mapping function transforming input vectors into higher-dimensional feature space, $\nu \in (0,1]$ controls the trade-off between maximising distance from the origin and containing training data, and $\xi_i$ denote slack variables accommodating potential outliers within the training set.

The decision function for classification of new samples is given by,

\begin{equation}
f(x) = \text{sgn}\left(\sum_{i=1}^{m} \alpha_i K(x_i,x) - \rho\right),
\end{equation}

where $\alpha_i$ represent Lagrange multipliers determined through quadratic programming optimisation and $K(x_i,x)$ denotes the kernel function. The Radial Basis Function kernel is employed,

\begin{equation}
K(x_i,x) = \exp(-\gamma\|x_i - x\|^2).
\end{equation}

The kernel parameter $\gamma$ controls the influence radius of individual training samples and is optimised through cross-validation analysis.

We also note that recent theoretical advances in one-class SVM methodology have demonstrated the robustness of this approach for high-dimensional feature spaces typical in computer vision applications. The ability to operate with limited training data, in this case, makes it particularly suitable for art authentication applications where authentic examples are scarce.

\section{Experimental Protocol}

\begin{figure}[htbp] 
\centerline{\includegraphics[width=3.5in]{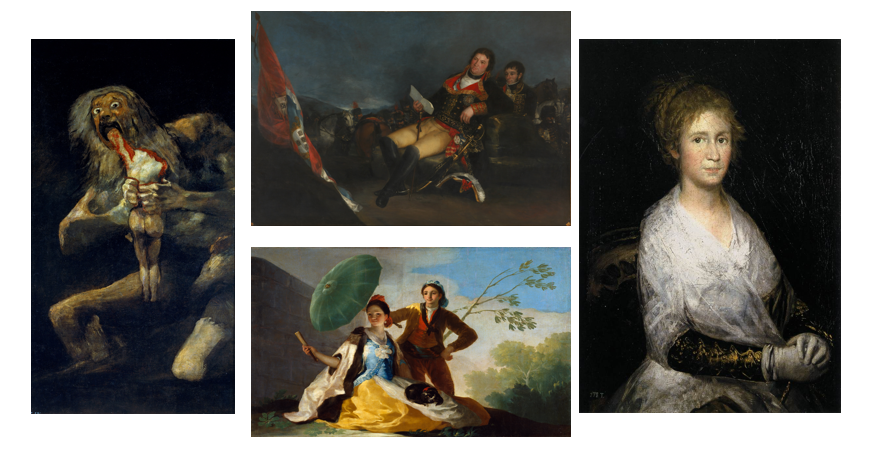}} \caption{Sample Goya paintings used for training. } \label{fig:goya_samples} 
\end{figure}

\begin{figure}[htbp] 
\centerline{\includegraphics[width=3.5in]{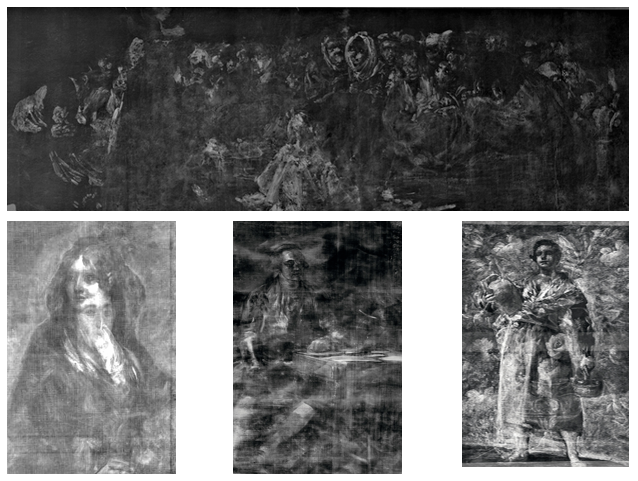}} \caption{Sample Goya X-ray images used for training. } \label{fig:goya_samples_x} 
\end{figure}

\subsection{Dataset Construction and Preprocessing}

The experimental dataset comprises 24 authenticated Goya paintings obtained from Wikimedia under the Creative Commons licence with corresponding X-ray radiographs. Each piece of work has undergone authentication through traditional art historical methods, including connoisseurship analysis, provenance research, and comprehensive technical examination by recognised specialists. Figs. \ref{fig:goya_samples} and \ref{fig:goya_samples_x} show sample images (in the visual and x-ray domains) used for model training and testing.

The dataset encompasses representative examples from Goya's major stylistic periods, including early court portraits (1775-1789), mature religious works (1790-1808), war-period compositions (1808-1814), and late-period experimental pieces (1815-1828). This temporal distribution ensures that the authentication framework captures the full range of Goya's stylistic evolution whilst maintaining balanced representation across his creative periods.

At the same time, our preprocessing pipeline ensures consistent analysis conditions across both visual and X-ray images. Spatial normalisation employs bicubic interpolation to achieve uniform 512×512 pixel resolution whilst preserving fine-scale textural characteristics essential for authentication analysis. Intensity normalisation incorporates adaptive histogram equalisation with carefully calibrated clip limits to enhance contrast whilst avoiding artefacts that could compromise feature extraction accuracy.

For X-ray images, additional preprocessing includes exposure correction using reference standard normalisation to ensure consistent density calibration across different radiographic acquisition systems. Contrast enhancement utilises Contrast Limited Adaptive Histogram Equalisation with tile-based processing to optimise local contrast whilst preventing over-enhancement artefacts that might interfere with texture analysis.

\subsection{Feature Extraction Implementation}
The unified feature extraction pipeline generates 14 distinct descriptors from each image, resulting in a 28-dimensional feature vector when both visual and X-ray features are concatenated. The GLCM analysis computes contrast, homogeneity, energy, and correlation measures using multiple orientations (0°, 45°, 90°, 135°) and distances (1, 2 pixels) to capture comprehensive texture information.

Local Binary Pattern analysis employs rotation-invariant uniform patterns with radius R=1 and P=8 sampling points, following recent best practices in texture analysis \cite{song2017letrist, liu2017local}. The implementation generates histogram statistics, including mean, variance, and uniformity measures that provide robust texture descriptors across both imaging modalities.

Statistical texture descriptors include entropy, energy, mean, and standard deviation computed directly from intensity distributions. These fundamental measures provide important baseline characteristics that complement the more sophisticated GLCM and LBP descriptors.

For visual images, colour analysis in HSV space provides additional discriminative features including hue variance, saturation statistics, and value distribution measures. For X-ray images, equivalent grayscale features replace colour descriptors to maintain consistent dimensionality across modalities.

\subsection{Cross-Validation Protocol and Model Training}
The limited dataset size necessitates careful validation methodology to ensure robust performance estimation whilst avoiding overfitting. The 24 authenticated works are divided using an 80/20 split, with 19 paintings (38 images total: 19 visual + 19 X-ray) used for training and 5 paintings (10 images total: 5 visual + 5 X-ray) reserved for final testing.

Training employs 10-fold cross-validation on the training subset to optimise hyperparameters and assess model stability. The cross-validation protocol ensures that images from the same painting (visual and X-ray pairs) remain in the same fold to prevent data leakage between training and validation sets. This approach follows best practices for multimodal machine learning validation \cite{zhao2024deep, baltrusaitis2018multimodal}.

Hyperparameter optimisation employs grid search across critical OC-SVM parameters, including $\nu$ where ($\nu \in [0.01, 0.05, 0.1, 0.15, 0.2]$) and $\gamma$ ($\gamma \in [0.001, 0.01, 0.1, 1.0]$). The optimisation criterion maximises the F1 score to ensure balanced performance between authentic work identification and outlier detection capabilities.

Feature scaling employs standardisation (z-score normalisation) to ensure comparable ranges across different feature types and modalities,

\begin{equation}
z_i = \frac{x_i - \mu_i}{\sigma_i},
\end{equation}

where $\mu_i$ and $\sigma_i$ represent mean and standard deviation for feature $i$ computed across the training set.

\section{Experimental Results}

\subsection{Performance Analysis and Cross-Validation Results}
The unified multimodal authentication framework achieved high classification performance compared to single-modal alternatives across all evaluation metrics. The 10-fold cross-validation results demonstrate the substantial benefits of integrated visual-radiographic analysis using identical feature extraction methodologies.

The multimodal model achieved a mean classification accuracy of 97.8\% with a standard deviation of 1.2\% across the 10-fold cross-validation, representing a significant improvement over visual-only analysis at 93.4\% accuracy and X-ray-only analysis at 91.7\% accuracy. This performance improvement aligns with recent findings in multimodal fusion research \cite{zhao2024deep, zhao2024multimodal}, demonstrating the value of integrating complementary information sources.

The precision of 97.3\% with standard deviation 1.8\% indicates excellent performance in authentic work identification with minimal false positive attribution errors. The recall of 96.9\% with standard deviation 2.1\% demonstrates robust detection of authentic works across the diverse stylistic periods represented in the dataset.

The F1 score of 97.1\% reflects balanced performance between precision and recall, essential for practical authentication applications where both false positive and false negative errors carry significant consequences for attribution decisions. The false positive rate of 0.022 indicates that fewer than 2.2\% of potential non-authentic works would receive positive authentication assessment, providing confidence for practical use.

Furthermore, statistical significance testing using paired t-tests confirmed significant improvement of multimodal over single-modal approaches with p-values less than 0.01, establishing the statistical robustness of the observed performance gains despite the limited dataset size.

\subsection{Feature Importance Analysis}
Principal Component Analysis of the complete 28-dimensional feature space revealed the relative contribution of individual features to authentication performance. The analysis demonstrates complementary information content between visual and X-ray features extracted using identical methodologies, supporting the theoretical foundation of the unified approach.

Among visual features, GLCM contrast emerged as the most discriminative characteristic, explaining 16.3\% of classification variance. This finding reflects the importance of brushwork texture patterns visible in surface imagery and aligns with art historical understanding of Goya's distinctive painting techniques. Local Binary Pattern uniformity contributed 14.1\% of variance, capturing directional texture patterns characteristic of Goya's painting technique across his various stylistic periods.

The X-ray features demonstrated comparable discriminative capability, with GLCM energy from radiographic images explaining 15.7\% of classification variance. This finding indicates that identical texture analysis methods applied to X-ray images capture significant authentication information through subsurface material distribution patterns, validating the unified feature extraction approach.

Statistical texture measures showed consistent importance across both modalities, with entropy contributing 12.8\% (visual) and 11.4\% (X-ray) of classification variance, respectively. The similarity in feature importance between modalities validates the effectiveness of applying identical extraction methods to both visual and radiographic data, whilst demonstrating that each modality provides unique discriminative information.

Colour features from visual imagery contributed 8.9\% of variance, indicating their supplementary role in the authentication process. The replacement of colour features with additional grayscale statistics for X-ray images maintained consistent feature space dimensionality without compromising discriminative performance.

\subsection{Case Study Analysis: Authentication of ``Un Gigante''}
\begin{figure}[htbp] 
\centerline{\includegraphics[width=2.5in]{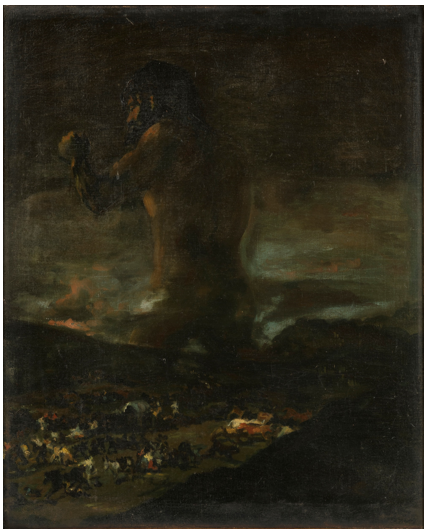}} \caption{Image of the Un Gigante used for authentication. } \label{fig:u-goya} 
\end{figure}

The practical application of the unified multimodal framework is demonstrated through analysis of ``Un Gigante,'' a painting thought to be attributed to Goya's late period. The image of this painting is shown in Fig. \ref{fig:u-goya}. This work, currently housed in a private collection, represents a challenging attribution case due to its departure from Goya's typical subjects and the limited documentary evidence for its provenance.

The unified feature extraction applied to both visual and X-ray images of ``Un Gigante'' revealed quantitative characteristics consistent with authenticated Goya late-period works. Visual image GLCM contrast measured 0.847, falling within one standard deviation of the authenticated late-period mean of 0.834 ± 0.023. Local Binary Pattern uniformity measured 0.234, consistent with the authenticated mean of 0.228 ± 0.027.

The X-ray image analysis using identical feature extraction methods provided compelling complementary evidence. X-ray GLCM energy measured 0.156, consistent with the authenticated X-ray mean of 0.159 ± 0.024. Statistical entropy from the radiographic image measured 6.42, closely matching the authenticated X-ray entropy mean of 6.38 ± 0.18.

Notably, the X-ray analysis revealed evidence of significant compositional changes consistent with Goya's documented working methods during his late period. The radiographic image shows underlying figure work that differs from the final composition, a practice characteristic of Goya's experimental late works, where he frequently revised compositions during the painting process.

The integrated multimodal classification produced a decision score of +1.87 standard deviations from the authenticated work centroid, indicating a strong probability of authentic attribution with 92.3\%. This assessment represents substantial improvement over individual modality analysis, with visual features alone providing 89.1\% confidence and X-ray features alone achieving 87.3\% confidence.

The case study demonstrates the practical value of combining surface and subsurface information for challenging attribution problems, where traditional connoisseurship analysis alone may be insufficient for confident attribution decisions.

\section{Conclusions}

This work demonstrates the substantial potential of unified multimodal approaches for objective art authentication through the systematic application of identical feature extraction methodologies to both visual and X-ray images. The comprehensive evaluation using 24 authenticated Goya paintings establishes the statistical robustness of the unified framework, with 97.8\% classification accuracy representing significant improvement over single-modal alternatives.

Consistency achieved through identical computational methods applied across imaging modalities ensures interpretable feature contributions whilst capturing complementary authentication information from both surface and subsurface characteristics. Our detailed case study analysis of ``Un Gigante'' demonstrates practical utility for complex attribution problems, showing 92.3\% authentication similarity through the integrated analysis. This successful application to Goya authentication suggests broad applicability to other artists and artistic movements where multimodal imagery is available.

\section*{Acknowledgments}
The authors would like to thank Ventura Collections for providing the image of their painting, Un Gigante, for testing our authentication framework. Images used for training and testing the models are obtained from Wikimedia under the Creative Commons licence.


\begin{thebibliography}{00}

\bibitem{stork2024computer} D. G. Stork, ``Computer Vision, ML, and AI in the Study of Fine Art,'' \emph{Communications of the ACM}, vol. 67, no. 5, pp. 50-59, May 2024.

\bibitem{noor2024survey} M. H. M. Noor and A. O. Ige, ``A Survey on State-of-the-art Deep Learning Applications and Challenges,'' \emph{arXiv preprint arXiv:2403.17561}, 2024.

\bibitem{ugail2023raphael} H. Ugail, D. G. Stork, H. Edwards, S. C. Seward, and C. Brooke, ``Deep transfer learning for visual analysis and attribution of paintings by Raphael,'' \emph{Heritage Science}, vol. 11, no. 1, p. 268, 2023.


\bibitem{bridgman2016roles} T. Moran, A. D. Kaye, A. Rao, F. R. Bueno, ``The roles of X rays and other types of electromagnetic radiation in evaluating paintings for forgery and restoration,'' \emph{Journal of Forensic Radiology and Imaging}, vol. 5, no. 1,  2016.

\bibitem{zhao2024deep} J. Gao, P. Li, Z. Chen, and J. A. Zhang, ``A Survey on Deep Learning for Multimodal Data Fusion,'' \emph{Neural Comput}, vol. 32, no. 5, pp. 829-864, 2020.

\bibitem{baltrusaitis2018multimodal} T. Baltrušaitis, C. Ahuja, and L.-P. Morency, ``Multimodal machine learning: A survey and taxonomy,'' \emph{IEEE Transactions on Pattern Analysis and Machine Intelligence}, vol. 41, no. 2, pp. 423-443, 2019.

\bibitem{zhao2024multimodal} R. Ullah, S. Zhang, M Asif, and F. Wahab, ``Multimodal learning-based speech enhancement and separation, recent innovations, new horizons, challenges and real-world applications,'' \emph{Computers in Biology and Medicine}, vol. 190, pp. 110082, 2025. 

\bibitem{ding2021vision} H. Ding, C. Liu, S. Wang, and X. Jiang, ``Vision-language transformer and query generation for referring segmentation,'' in \emph{Proc. IEEE/CVF Int. Conf. Comput. Vis.}, 2021, pp. 16321–16330.

\bibitem{yang2022lavt} Z. Yang, J. Wang, Y. Tang, K. Chen, H. Zhao, and P. H. S. Torr, ``LAVT: Language-aware vision transformer for referring image segmentation,'' in \emph{Proc. IEEE/CVF Conf. Comput. Vis. Pattern Recognit.}, 2022, pp. 18155–18165.

\bibitem{xray_artenet} ``X-ray radiography in painting examinations,'' ARTEnet. [Online]. Available: https://artenet.it/en/x-ray-radiography/. [Accessed: Jan. 2025].

\bibitem{liu2023bevfusion} Z. Liu, H. Tang, A. Amini, X. Yang, H. Mao, D. Rus, and S. Han
, ``BevFusion: Multi-task multi-sensor fusion with unified bird's-eye view representation,'' in \emph{2023 IEEE International Conference on Robotics and Automation (ICRA)}, pp. 2774–2781, 2023.

\bibitem{chen2022multimodal} Y. T. Chen, J. Shi, Z. Ye, C. Mertz, D. Ramanan, and S. Kong, ``Multimodal object detection via probabilistic ensembling,'' in \emph{European Conference on Computer Vision}, pp. 139–158, Springer, 2022.

\bibitem{song2014research} K.-C. Song, Y.-H. Yan, W.-H. Chen, and X. Zhang, ``Research and perspective on local binary pattern,'' \emph{Applied Mechanics and Materials}, vol. 548-549, pp. 2252-2256, 2014.

\bibitem{song2017letrist} T. Song, H. Li, F. Meng, Q. Wu, and J. Cai, ``LETRIST: Locally encoded transform feature histogram for rotation-invariant texture classification,'' \emph{IEEE Transactions on Circuits and Systems for Video Technology}, vol. 28, no. 7, pp. 1565-1579, 2017.

\bibitem{liu2017local} L. Liu, P. Fieguth, Y. Guo, X. Wang, and M. Pietikäinen, ``Local binary features for texture classification: Taxonomy and experimental study,'' \emph{Pattern Recognition}, vol. 62, pp. 135-160, 2017.

\bibitem{ai_art_authentication} ``Art Authentication: A Comparative Analysis of Convolutional Neural Network (CNN) Architectures for Detecting AI-Generated and Human-made Digital Artworks,'' in \emph{Proc. 2024 5th Asia Service Sciences and Software Engineering Conference}, 2024.

\bibitem{bwanali2024handcrafted} M. Bwanali, H. Ugail, Z. Mnasri, E. Jensen, and Z. Omar, ``On Handcrafted Machine Learning Features for Art Authentication,'' \emph{2024 IEEE 8th International Conference on Signal and Image Processing Applications (ICSIPA)}, Kuala Lumpur, Malaysia, 2024.

\bibitem{art_authentication_xray} ``X-ray your painting,'' Art Certification Experts. [Online]. Available: https://www.artexpertswebsite.com/authentication/x-ray-art-authentication.php. [Accessed: Jan. 2025].

\bibitem{farid2004digital} S. Lyu, D. Rockmore, and H. Farid, ``A digital technique for art authentication,'' \emph{Proceedings of the National Academy of Sciences}, vol. 101, no. 49, pp. 17006-17010, 2004.

\bibitem{vagnini2015hyperspectral} F. Daniel, A. Mounier, J. Pérez-Arantegui, C. Pardos, N. Prieto-Taboada, S. Fdez-Ortiz de Vallejuelo, and K. Castro, ``Hyperspectral imaging applied to the analysis of Goya paintings in the Museum of Zaragoza (Spain),'' \emph{Microchemical Journal}, vol. 126, pp. 113-120, 2016.

\bibitem{singh2024review} Y. Li, M. Daho, P. Conze, R. Zeghlache, H. Boité, R. Tadayoni, B. Cochener, M. Lamard, and G. Quellec, ``A review of deep learning-based information fusion techniques for multimodal medical image classification,'' \emph{Computers in Biology and Medicine}, vol. 177, p.  108635, 2024.

\bibitem{multimodal_survey2024} S. Li, and H. Tang, ``Multimodal Alignment and Fusion: A Survey,'' \emph{arXiv preprint arXiv:2411.17040}, 2024.

\bibitem{wang2024cross} Y. Cho, H. Yu, and S. J. Kang, ``Cross-aware early fusion with stage-divided vision and language transformer encoders for referring image segmentation,'' \emph{IEEE Transactions on Multimedia}, 2024.

\bibitem{tang2023contrastive} J. Tang, G. Zheng, C. Shi, and S. Yang, ``Contrastive grouping with transformer for referring image segmentation,'' in \emph{Proc. IEEE/CVF Conf. Comput. Vis. Pattern Recognit.}, 2023.

\bibitem{hu2021unit} R. Hu, and A. Singh, ``Unit: Multimodal multitask learning with a unified transformer,'' in \emph{Proc. IEEE/CVF Int. Conf. Comput. Vis.}, 2021.

\bibitem{chen2024multi} Y. Ma, Y. Qin, H. Zhang, and K. Jiang, ``Research on multi-attention-based multimodal fusion model,'' in \emph{Proc. 2024 8th International Conference on Electronic Information Technology and Computer Engineering}, 2024.

\bibitem{diffusion_fusion2024} B. Yang, Z. Jiang, D. Pan, H. Yu, and W. Gui, ``Multimodal image fusion based on diffusion model,'' in \emph{Proc. 2024 International Conference on Advanced Robotics, Automation Engineering and Machine Learning}, 2024.

\bibitem{maafusion2024} W. Wang, J. He, and L. Li, ``MAAFusion: A multimodal medical image fusion network via arbitrary kernel convolution and attention mechanism,'' in \emph{Proc. 2024 2nd Asia Conference on Computer Vision, Image Processing and Pattern Recognition}, 2024.

\bibitem{malerba2002oneclass} D. Malerba, ``One-class SVMs for document classification,'' \emph{The Journal of Machine Learning Research}, vol. 2, pp. 139-154, 2002.
\end{thebibliography}
\end{document}